# n-Valued Refined Neutrosophic Logic and Its Applications to Physics

Florentin Smarandache, Ph. D.
University of New Mexico
Math & Science Division
705 Gurley Ave.
Gallup, NM 87301, USA
E-mail:smarand@unm.edu

**Abstract.**

In this paper we present a short history of logics: from particular cases of *2*-symbol or numerical valued logic to the general case of *n*-symbol or numerical valued logic. We show generalizations of *2*-valued Boolean logic to fuzzy logic, also from the Kleene's and Lukasiewicz' *3*-symbol valued logics or Belnap's *4*-symbol valued logic to the most general *n-symbol or numerical valued refined neutrosophic logic*. Two classes of neutrosophic norm (*n-norm*) and neutrosophic conorm (*n-conorm*) are defined. Examples of applications of neutrosophic logic to physics are listed in the last section.
Similar generalizations can be done for *n-Valued Refined Neutrosophic Set*, and respectively *n-Valued Refined Neutrosopjhic Probability*.

## 1. Two-Valued Logic

a) The Two Symbol-Valued Logic.

It is the Chinese philosophy: *Yin* and *Yang* (or Femininity and Masculinity) as contraries:

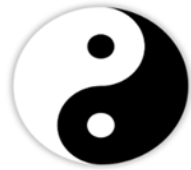

*Fig 1. Ying and Yang*

It is also the Classical or *Boolean Logic*, which has two symbol-values: truth *T* and falsity *F*.

b) The Two Numerical-Valued Logic.
   It is also the Classical or *Boolean Logic*, which has two numerical-values: truth *1* and falsity *0*.
   More general it is the *Fuzzy Logic*, where the truth (*T*) and the falsity (*F*) can be any numbers in *[0,1]* such that *T + F = 1*.
Even more general, *T* and *F* can be subsets of *[0, 1]*.

## 2. Three-Valued Logic

a) The <u>Three Symbol-Valued Logics</u>:

i) *Łukasiewicz 's Logic*: True, False, and Possible.

ii) *Kleene's Logic*: True, False, Unknown (or Undefined).

iii) Chinese philosophy extended to: *Yin, Yang,* and *Neuter* (or Femininity, Masculinity, and Neutrality) – as in Neutrosophy.

Neutrosophy philosophy was born from neutrality between various philosophies. <u>Connected with Extenics</u> (Prof. Cai Wen, 1983), and Paradoxism (F. Smarandache, 1980).

<u>Neutrosophy</u> is a new branch of philosophy that studies the origin, nature, and scope of neutralities, as well as their interactions with different ideational spectra.

This theory considers every notion or idea <A> together with its opposite or negation <antiA> and with their spectrum of neutralities <neutA> in between them (i.e. notions or ideas supporting neither <A> nor <antiA>).

The <neutA> and <antiA> ideas together are referred to as <nonA>.

Neutrosophy is a generalization of Hegel's dialectics (the last one is based on <A> and <antiA> only).

According to this theory every idea <A> tends to be neutralized and balanced by <antiA> and <nonA> ideas - as a state of equilibrium.

In a classical way <A>, <neutA>, <antiA> are disjoint two by two. But, since in many cases the borders between notions are vague, imprecise, Sorites, it is possible that <A>, <neutA>, <antiA> (and <nonA> of course) have common parts two by two, or even all three of them as well. <u>Such contradictions involves Extenics.</u> Neutrosophy is the base of all neutrosophics and it is used in engineering applications (especially for software and information fusion), medicine, military, airspace, cybernetics, physics.

b) The <u>Three Numerical-Valued Logic</u>:

*i) Kleene's Logic*: True (*1*), False (*0*), Unknown (or Undefined) (*1/2*), and uses "min" for ∧, "max" for ∨, and "1-" for negation.

ii) More general is the *Neutrosophic Logic* [Smarandache, 1995], where the truth (*T*) and the falsity (*F*) and the indeterminacy (*I*) can be any numbers in *[0, 1],* then $0 \leq T + I + F \leq 3$.

More general: Truth (*T*), Falsity (*F*), and Indeterminacy (*I*) are standard or nonstandard subsets of the nonstandard interval *]-0, 1+[*.

### 3. Four-Valued Logic

a) The <u>Four Symbol-Valued Logic</u>

i) It is *Belnap's Logic:* True (*T*), False (*F*), Unknown (*U*), and Contradiction (*C*), where *T, F, U, C* are symbols, not numbers.

Below is the Belnap's conjunction operator table:

| ∩ | F | U | C | T |
|---|---|---|---|---|
| **F** | **F** | **F** | **F** | **F** |
| **U** | **F** | **U** | **F** | **U** |
| **C** | **F** | **F** | **C** | **C** |
| **T** | **F** | **U** | **C** | **T** |

Restricted to *T, F, U,* and to *T, F, C,* the Belnap connectives coincide with the connectives in Kleene's logic.

ii) Let *G* = Ignorance. We can also propose the following two 4-Symbol Valued Logics: *(T, F, U, G),* and *(T, F, C, G).*

iii) *Absolute-Relative 2-, 3-, 4-, 5-, or 6-Symbol Valued Logics* [Smarandache, 1995].
Let $T_A$ be truth in all possible worlds (according to Leibniz's definition);
$T_R$ be truth in at last one world but not in all worlds;
and similarly let $I_A$ be indeterminacy in all possible worlds;
$I_R$ be indeterminacy in at last one world but not in all worlds;
also let $F_A$ be falsity in all possible worlds;
$F_R$ be falsity in at last one world but not in all worlds;

    Then we can form several Absolute-Relative 2-, 3-, 4-, 5-, or 6-Symbol Valued Logics just taking combinations of the symbols $T_A$, $T_R$, $I_A$, $I_R$, $F_A$, and $F_R$.

    As particular cases, very interesting would be to study the Absolute-Relative 4-Symbol Valued Logic *($T_A$, $T_R$, $F_A$, $F_R$),* as well as the Absolute-Relative *6*-Symbol Valued Logic *($T_A$, $T_R$, $I_A$, $I_R$, $F_A$, $F_R$).*

   b) *Four Numerical-Valued Neutrosophic Logic:* Indeterminacy I is refined (split) as U = Unknown, and C = contradiction.
   T, F, U, C are subsets of [0, 1], instead of symbols;
   This logic generalizes Belnap's logic since one gets a degree of truth, a degree of falsity, a degree of unknown, and a degree of contradiction.
   Since C = T∧F, this logic involves the Extenics.

## 4. Five-Valued Logic

   *a)   Five Symbol-Valued Neutrosophic Logic* [Smarandache, 1995]:
   Indeterminacy *I* is refined (split) as *U* = Unknown, *C* = contradiction, and *G* = ignorance; where the symbols represent:
*T* = truth;
*F* = falsity;
*U* = neither *T* nor *F* (undefined);
*C* = *T∧F,* which involves the Extenics;
*G* = *T∨F.*

   *b)   If T, F, U, C, G are subsets of [0, 1] then we get: a Five Numerical-Valued Neutrosophic Logic.*

## 5. Seven-Valued Logic

   *a)   Seven Symbol-Valued Neutrosophic Logic* [Smarandache, 1995]:
   I is refined (split) as *U, C, G,* but *T* also is refined as $T_A$ = absolute truth and $T_R$ = relative truth, and *F* is refined as $F_A$ = absolute falsity and $F_R$ = relative falsity. Where:
*U* = neither (*$T_A$ or $T_R$*) nor (*$F_A$ or $F_R$*) (i.e. undefined);

$C = (T_A \text{ or } T_R) \wedge (F_A \text{ or } F_R)$ (i.e. Contradiction), which involves the Extenics;
$G = (T_A \text{ or } T_R) \vee (F_A \text{ or } F_R)$ (i.e. Ignorance).
All are symbols.

    *b)* But if $T_A, T_R, F_A, F_R, U, C, G$ are subsets of *[0, 1]*, then we get a *Seven Numerical-Valued Neutrosophic Logic*.

## 6. **n-Valued Logic**

    *a) The n-Symbol-Valued Refined Neutrosophic Logic* [Smarandache, 1995].
In general:
*T* can be split into many types of truths: $T_1, T_2, ..., T_p$, and *I* into many types of indeterminacies: $I_1, I_2, ..., I_r$, and *F* into many types of falsities: $F_1, F_2, ..., F_s$, where all *p, r, s ≥ 1* are integers, and *p + r + s = n*. Even more: T, I, and/or F (or any of their subcomponents $T_j$, $I_k$, and/or $F_l$) can be countable or uncountable infinite sets.
All subcomponents $T_j, I_k, F_l$ are symbols for *j∈ {1,2,...,p}, k∈ {1,2,...,r}*, and *l∈ {1,2,...,s}*.
If at least one $I_k = T_j \wedge F_l$ = contradiction, we get again the Extenics.

    b) *The n-Numerical-Valued Refined Neutrosophic Logic.*
In the same way, but all subcomponents $T_j, I_k, F_l$ are not symbols, but subsets of *]⁻0, 1⁺[*, for all *j ∈ {1,2,...,p}*, all *k ∈ {1,2,...,r}*, and all *l ∈ {1,2,...,s}*. *Even more: T, I, and/or F (or any of their subcomponents $T_j$ ,$I_k$, and/or $F_l$) can be countable or uncountable infinite sets.*
If all sources of information that separately provide neutrosophic values for a specific subcomponent are independent sources, then in the general case we consider that each of the subcomponents $T_j, I_k, F_l$ is independent with respect to the others and it is in the non-standard set *]⁻0, 1⁺[*. Therefore per total we have for crisp neutrosophic value subcomponents $T_j, I_k, F_l$ that:

$$^{-}0 \leq \sum_{j=1}^{p} T_j + \sum_{k=1}^{r} I_k + \sum_{l=1}^{s} F_l \leq n^{+} \qquad (1)$$

where of course *n = p + r + s* as above.

If there are some dependent sources (or respectively some dependent subcomponents), we can treat those dependent subcomponents together. For example, if $T_2$ and $I_3$ are dependent, we put them together as $^{-}0 \leq T_2 + I_3 \leq 1^{+}$.
The non-standard unit interval *]⁻0, 1⁺[* , used to make a distinction between *absolute* and *relative* truth/indeterminacy/falsehood in philosophical applications, is replaced for simplicity with the standard (classical) unit interval *[0, 1]* for technical applications.

For at least one $I_k = T_j \wedge F_l$ = contradiction, we get again the Extenics.

## 7. **n-Valued Neutrosophic Logic Connectors**

### a) **n-Norm and n-Conorm defined on combinations of t-Norm and t-Conorm**

The n-norm is actually the neutrosophic conjunction operator, NEUTROSOPHIC AND ($\wedge_n$); while the n-conorm is the neutrosophic disjunction operator, NEUTROSOPHIC OR ($\vee_n$).

One can use the t-norm and t-conorm operators from the fuzzy logic in order to define the **n-norm** and respectively **n-conorm** in neutrosophic logic:

$$n\text{-}norm(\ (T_j)_{j=\{1,2,...,p\}},\ (I_k)_{k=\{1,2,...,r\}},\ (F_l)_{l=\{1,2,...,s\}}\ ) = \qquad\qquad (2)$$

$$(\ [t\text{-}norm(T_j)]_{j=\{1,2,...,p\}},\ [t\text{-}conorm(I_k)]_{k=\{1,2,...,r\}},\ [t\text{-}conorm(F_l)]_{l=\{1,2,...,s\}}\ )$$

and

$$n\text{-}conorm(\ (T_j)_{j=\{1,2,...,p\}},\ (I_k)_{k=\{1,2,...,r\}},\ (F_l)_{l=\{1,2,...,s\}}\ ) = \qquad\qquad (3)$$

$$(\ [t\text{-}conorm(T_j)]_{j=\{1,2,...,p\}},\ [t\text{-}norm(I_k)]_{k=\{1,2,...,r\}},\ [t\text{-}norm(F_l)]_{l=\{1,2,...,s\}}\ )$$

and then one normalizes if needed.

Since the n-norms/n-conorms, alike t-norms/t-conorms, can only approximate the inter-connectivity between two n-Valued Neutrosophic Propositions, there are many versions of these approximations.

For example, for the n-norm:

the indeterminate (sub)components $I_k$ alone can be combined with the t-conorm in a pessimistic way [i.e. lower bound], or with the t-norm in an optimistic way [upper bound];

while for the n-conorm:

the indeterminate (sub)components $I_k$ alone can be combined with the t-norm in a pessimistic way [i.e. lower bound], or with the t-conorm in an optimistic way [upper bound].

In general, if one uses in defining an n-norm/n-conorm for example the t-norm $min\{x,\ y\}$ then it is indicated that the corresponding t-conorm used be $max\{x,\ y\}$;  or if the t-norm used is the product $x{\cdot}y$ then the corresponding t-conorm should be $x+y-x{\cdot}y$;  and similarly if the t-norm used is $max\{0,\ x+y-1\}$ then the corresponding t-conorm should be $min\{x+y,\ 1\}$;  and so on.

Yet, it is still possible to define the n-norm and n-conorm using different types of t-norms and t-conorms.

**b)  N-norm and n-conorm based on priorities.**
For the n-norm we can consider the priority: T < I < F, where the subcomponents are supposed to conform with similar priorities, i.e.

$$T_1 < T_2 < ... < T_p < I_1 < I_2 < ... < I_r < F_1 < F_2 < ... < F_s. \qquad\qquad (4)$$

While for the n-conorm one has the opposite priorities: T > I > F, or for the refined case:

$$T_1 > T_2 > ... > T_p > I_1 > I_2 > ... > I_r > F_1 > F_2 > ... > F_s. \qquad\qquad (5)$$

By definition A < B means that all products between A and B go to B (the bigger).

Let's say, one has two neutrosophic values in simple (non-refined case):

$(T_x, I_x, F_x)$                                                             (6)

and

$(T_y, I_y, F_y)$.                                                            (7)

Applying the n-norm to both of them, with priorities $T < I < F$, we get:

$(T_x, I_x, F_x) \wedge_n (T_y, I_y, F_y) = (T_xT_y, T_xI_y + T_yI_x + I_xI_y, T_xF_y + T_yF_x + I_xF_y + I_yF_x + F_xF_y)$    (8)

Applying the n-conorm to both of them, with priorities $T > I > F$, we get:

$(T_x, I_x, F_x) \vee_n (T_y, I_y, F_y) = (T_xT_y + T_xI_y + T_yI_x + T_xF_y + T_yF_x, I_xI_y + I_xF_y + I_yF_x, F_xF_y)$.    (9)

In a lower bound (pessimistic) n-norm one considers the priorities $T < I < F$, while in an upper bound (optimistic) n-norm one considers the priorities $I < T < F$.

Whereas, in an upper bound (optimistic) n-conorm one considers $T > I > F$, while in a lower bound (pessimistic) n-conorm one considers the priorities $T > F > I$.

Various priorities can be employed by other researchers depending on each particular application.

### 8. Particular Cases

If in *6 a)* and *b)* one has all $I_k = 0$, $k = \{1,2,...,r\}$, we get the **n-Valued Refined Fuzzy Logic**.

If in *6 a)* and *b)* one has only one type of indeterminacy, i.e. $k = 1$, hence $I_1 = I > 0$, we get the **n-Valued Refined Intuitionistic Fuzzy Logic**.

### 9. Distinction between Neutrosophic Physics and Paradoxist Physics

Firstly, we make a distinction between Neutrosophic Physics and Paradoxist Physics.

#### a) Neutrosophic Physics.

Let <A> be a physical entity (i.e. concept, notion, object, space, field, idea, law, property, state, attribute, theorem, theory, etc.), <antiA> be the opposite of <A>, and <neutA> be their neutral (i.e. neither <A> nor <antiA>, but in between).

Neutrosophic Physics is a mixture of two or three of these entities <A>, <antiA>, and <neutA> that hold together.

Therefore, we can have neutrosophic fields, and neutrosophic objects, neutrosophic states, etc.

#### b) Paradoxist Physics.

Neutrosophic Physics is an extension of Paradoxist Physics, since Paradoxist Physics is a combination of physical contradictories *<A>* and *<antiA>* only that hold together, without referring to their neutrality *<neutA>*. Paradoxist Physics describes collections of objects or states that are individually characterized by contradictory properties, or are characterized neither by a property nor by the opposite of that property, or are composed of contradictory sub-elements. Such objects or states are called *paradoxist entities*.

These domains of research were set up in the *1995* within the frame of neutrosophy, neutrosophic logic/set/probability/statistics.

## 10. n-Valued Refined Neutrosophic Logic Applied to Physics

There are many cases in the scientific (and also in humanistic) fields that two or three of these items *<A>*, *<antiA>*, and *<neutA>* simultaneously coexist.

Several **Examples** of paradoxist and neutrosophic entities:

- anions in two spatial dimensions are arbitrary spin particles that are neither bosons (integer spin) nor fermions (half integer spin);

- among possible *Dark Matter* candidates there may be exotic particles that are neither *Dirac* nor *Majorana fermions;*

- mercury (Hg) is a state that is neither liquid nor solid under normal conditions at room temperature;

- non-magnetic materials are neither ferromagnetic nor anti-ferromagnetic;

- quark gluon plasma (QGP) is a phase formed by quasi-free quarks and gluons that behaves neither like a conventional plasma nor as an ordinary liquid;

- *unmatter,* which is formed by matter and antimatter that bind together (F. Smarandache, 2004);

- neutral kaon, which is a pion & anti-pion composite (R. M. Santilli, 1978) and thus a form of unmatter;

- neutrosophic methods in General Relativity (D. Rabounski, F. Smarandache, L. Borissova, 2005);

- neutrosophic cosmological model (D. Rabounski, L. Borissova, 2011);

- neutrosophic gravitation (D. Rabounski);

- qubit and generally quantum superposition of states;

- semiconductors are neither conductors nor isolators;

- semi-transparent optical components are neither opaque nor perfectly transparent to light;

- quantum states are metastable (neither perfectly stable, nor unstable);

- neutrino-photon doublet (E. Goldfain);

- the "multiplet" of elementary particles is a kind of 'neutrosophic field' with two or more values (E. Goldfain, 2011);

- A "neutrosophic field" can be generalized to that of operators whose action is selective. The effect of the neutrosophic field is somehow equivalent with the "tunneling" from the solid physics, or with the "spontaneous symmetry breaking" (SSB) where there is an internal symmetry which is broken by a particular selection of the vacuum state (E. Goldfain).

Etc.

**Conclusion**

Many types of logics have been presented above. For the most general logic, the n-valued refined neutrosophic logic, we presented two classes of neutrosophic operators to be used in combinations of neutrosophic valued propositions in physics.

Similar generalizations are done for **n-Valued Refined Neutrosophic Set**, and respectively **n-Valued Refined Neutrosopjhic Probability**.